\pdfoutput=1
\documentclass[11pt]{article}

\usepackage{EMNLP2023}

\usepackage{times}
\usepackage{latexsym}

\usepackage[T1]{fontenc}
\usepackage[utf8]{inputenc}

\usepackage{microtype}

\usepackage{inconsolata}
\usepackage{amsmath}
\usepackage{graphicx,subcaption}
\usepackage{CJK}
\usepackage{multirow}
\usepackage{booktabs}

\usepackage{algorithm}
\usepackage[noend]{algpseudocode}
\usepackage{enumitem}
\usepackage{placeins}
\usepackage{cancel} % add by wjn

\definecolor{gray-green}{HTML}{4CCBCD}

\newlist{noitemize}{itemize}{1}

\newcommand{\ignore}[1]{}

\title{Prompting Large Language Models with Chain-of-Thought for Few-Shot Knowledge Base Question Generation}

\author{
     Yuanyuan Liang$^1$, Jianing Wang$^1$, Hanlun Zhu$^1$,  Lei Wang$^2$ \\
      \textbf{Weining Qian$^1$, Yunshi Lan$^1$\thanks{*Corresponding author}}\\
  $^1$ East China Normal University, $^2$ Singapore Management University\\
  	leonyuany@stu.ecnu.edu.cn, \{lygwjn, timberflowing\}@gmail.com \\
 lei.wang.2019@phdcs.smu.edu.sg, \{wnqian, yslan\}@dase.ecnu.edu.cn 
   }

\begin{document}
\maketitle
\begin{abstract}
The task of Question Generation over Knowledge Bases (KBQG) aims to convert a logical form into a natural language question. 
For the sake of expensive cost of large-scale question annotation, the methods of KBQG under low-resource scenarios urgently need to be developed.
However, current methods heavily rely on annotated data for fine-tuning, which is not well-suited for few-shot question generation. 
%\leicomment{The reason sounds unnatural. why the few-shot scenario is necessary or important.}
The emergence of Large Language Models (LLMs) has shown their impressive generalization ability in few-shot tasks.
Inspired by Chain-of-Thought (CoT) prompting, which is an in-context learning strategy for reasoning, we formulate KBQG task as a reasoning problem, where the generation of a complete question is split into a series of sub-question generation.
Our proposed prompting method KQG-CoT first selects supportive logical forms from the unlabeled data pool taking account of the characteristics of the logical form.
% \jncomment{Then, we aim to select multiple exemplars construct a task-specific prompt to elicit the LLM to generate}
Then, we construct a task-specific prompt to guide LLMs to generate complicated questions based on selective logic forms.
% \leicomment{I am not sure the use of elicit here is correct. elicit sth. from someone. e.g., elicit reasoning ability from LLMs}
To further ensure prompt quality, we extend KQG-CoT into KQG-CoT+ via sorting the logical forms by their complexity.
We conduct extensive experiments over three public KBQG datasets.
The results demonstrate that our prompting method consistently outperforms other prompting baselines on the evaluated datasets. 
Remarkably, our KQG-CoT+ method could surpass existing few-shot SoTA results of the PathQuestions dataset by $18.25$, $10.72$, and $10.18$ absolute points on BLEU-4, METEOR, and ROUGE-L, respectively.

%and its  has proven effective in enhancing LLM performance for complex tasks by providing an intermediate reasoning chain that leads to the final output.
%The KBQA task can also be considered a reasoning problem, where the inference of corresponding phrases is based on logical form reasoning, and the logical form is continuously expanded to obtain the final natural language question.Therefore, we propose a method for few-shot KBQA tasks that with CoT prompting to enhance performance of language models. To construct an effective CoT prompt, we employ a two-step process. The first step is to retrieve Supportive Logical Forms. We begin by obtaining the skeleton of logical forms, followed by clustering these skeletons.We sample at least one skeleton from the clusters and rearrange the selected skeletons as needed.In the second step, we identify the logical forms that correspond to the selected skeletons and construct the CoT prompt accordingly.
%We conducted experiments on three KBQG datasets: WebQuestions, PathQuestions, and GrailQA. The results demonstrate the effectiveness of our proposed method.

\end{abstract}

\section{Introduction}
\label{sec:intro}

\begin{figure} 
\centering
\includegraphics[width=0.47\textwidth]{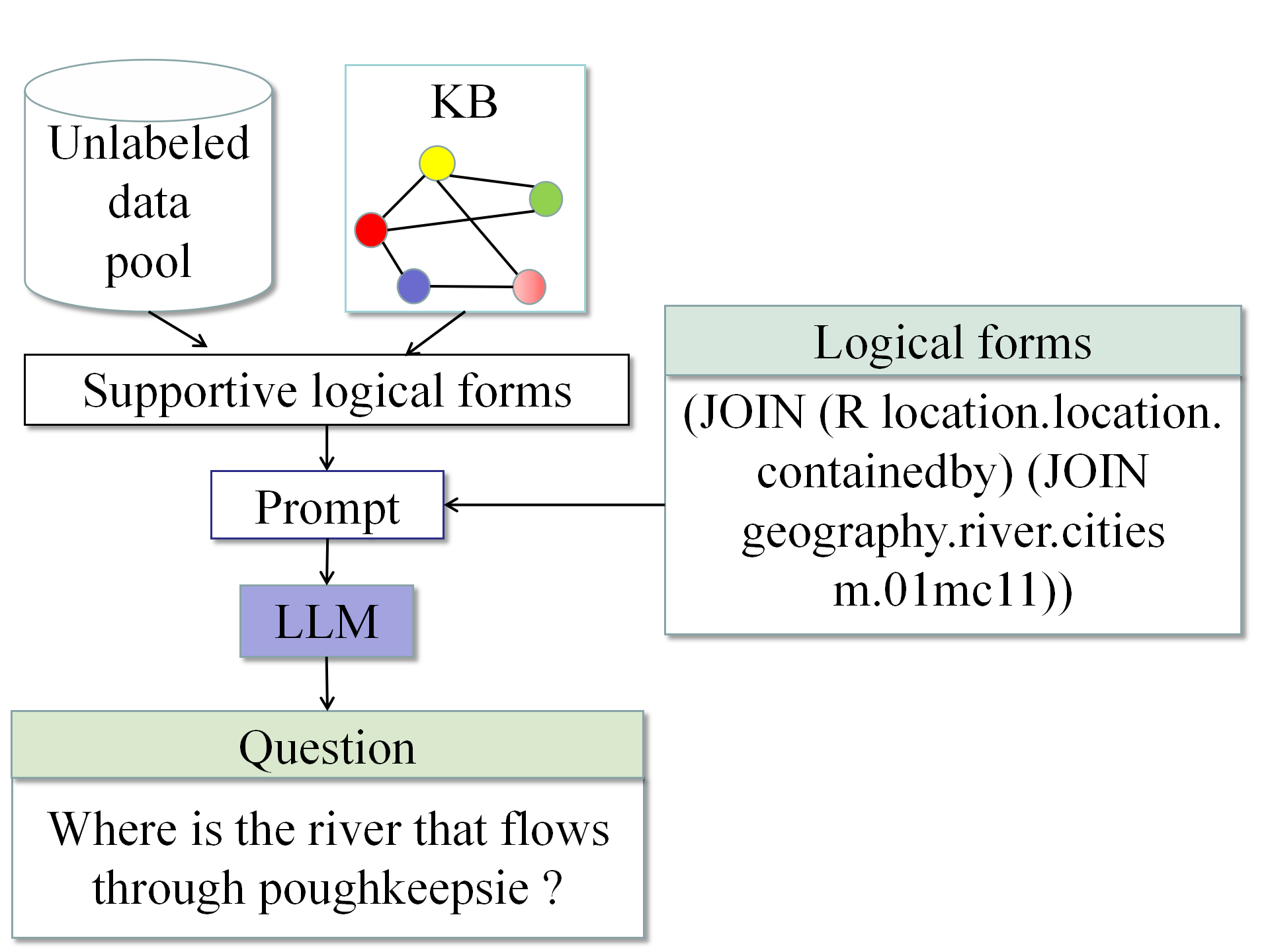}
% \vspace{-0.5cm}
\caption{Overview of KQG-CoT framework.
% \yscomment{How about we call ``supportive logical forms'' and ``prompts'' instead of ``supportive demonstrations'' and ``chain prompts''.}
% \yscomment{Please make the first letter of the question uppercase.}
}
\label{fig:kbqg_pipeline}
\end{figure}

%  一、简单介绍KBQG的用途
Question generation task requires a system to produce natural language questions based on the given context. 
% When the given context is a logical form deriving from Knowledge Bases (KBs), it becomes KBQG~\cite{guo-etal-2022:dsm}. 
KBQG~\cite{guo-etal-2022:dsm} is one of the imperative question generation tasks when the given context derived from Knowledge Bases (KBs) is in the form of logical.
KBQG has attracted increasing interests from both the industry and academia due to its potential for data augmentation in QA systems~\cite{xiong-2022:AutoQGS,Chen-2020:TowardSG} and its ability to assist dialogue systems in creating coherent questions~\cite{lee-IEEE-2018:Automatic_QG}.

Existing studies~\cite{Kumar-ISWC-2019:MHQG,ke-etal-2021:jointgt,fei-etal-2022:lfkqg,guo-etal-2022:dsm,Chen-2020:TowardSG} for KBQG tasks have predominantly utilized neural network-based approaches and demonstrated impressive performance by conducting fine-tuning on extensive training datasets.
However, as the collection of KBQG data is labor-intensive, researchers start paying attention to the few-shot KBQG tasks~\cite{xiong-2022:AutoQGS}, where a great challenge is posed for suppliers with limited resources: 
1) A great deal of annotated data is demanded to allow the existing fine-tuned models to generalize well over different logical forms. 
%But the conventional methods are developed based on full supervision.
However, due to the limitations of low-resource availability, training conventional models by fine-tuning on the full data becomes unrealistic.
% \leicomment{full supervision might be agnostic to number of data examples.}
2) A logical form is composed of entities, relations, and query grammar. Having logical forms with various combinations of these basic components is crucial to uphold the model's capability for compositional generalization.
% \jncomment{Logical forms are prerequisites to maintain the compositional generalization capability of a model, which mainly consists of multiple entities, relations, and corresponding query grammar}
The lack of data leads to a compositional challenge to the KBQG tasks~\cite{Gu-2020:GrailQA}. 
% 3) Some logical forms are complicated due to the aggregation, superlative, and comparative operations included.
% Developing a KBQG method with rich expressions is difficult, especially under a low-resource scenario.
3) Certain logical forms can become complex when operations such as aggregation, superlatives, and comparisons are involved. Representing these logical forms presents additional challenges.
Moreover, developing a KBQG method that incorporates diverse and elaborate expressions becomes particularly difficult in such low-resource scenarios~\cite{xiong-2022:AutoQGS,guo-etal-2022:dsm}. 

% \leicomment{Are these challenges in low-resource settings directly connected to advantages of LLM-based methods?}

Recently, LLMs such as GPT-3 and Codex~\cite{gao:arxiv2022,suzgun:arxiv2022,wei-2022:chain,wang:arxiv2023} have proven their strong generalizability on a wide range of  few-shot and zero-shot tasks with CoT, including text interpretation, computer vision, planning and reasoning. 
%\leicomment{The term in-context learning = few-shot learning in many cases. I am afraid ``zero-shot in-context learning'' may not be an appropriate description. (although there is a work named zero-shot icl with pseudo demos.}
Meanwhile, a line of work~\cite{kasner:arxiv2022,moiseev-etal-2022:skill,Andrus-2022-Enhanced_Story,trajanoska-2023:enhancing,xie2023empirical}
% ~\yscomment{Please help me double check the citation whether they are taking structural sequence as input.} 
validates that LLMs have the strong capability to accurately capture the semantics of relations between values in the data, enabling to transform the structured instructions to narrative text.
The above studies inspire us to explore few-shot KBQG tasks by prompting LLMs with CoT. 
 % \leicomment{We should underline unprecedented few-shot and zero-shot ability of LLMs, leading to our research of exploring low-resource KBQG.}

% However, how to apply LLMs to KBQG with in-context learning is still unclear.
% On the one hand, unlike code generation or question answering tasks, where the input is self-contained narrative, KBQG has the input with KB-specific items. So it is crucial to arrange the input to LLMs in the format of an easy-to-understand verbalizer with the awareness of the schema of the KB.
% On the other hand, in low-resource scenarios, it is impossible that the system can get access to numerous annotated data pairs.
% We should make full use of a collection of unlabeled logical forms.
% \jncomment{A natural question arise: ``how to better elicit LLMs on few-shot KBQG by designing chain-of-thought prompting?''}

However, how to apply LLMs to KBQG with CoT is still unclear. On one hand, KBQG differs from tasks like code generation or question answering, as it involves incorporating KB-specific items into the input instead of self-contained narratives. Therefore, formatting the input in an easily understandable manner while considering the KB schema is crucial. On the other hand, the challenge lies in designing effective CoT prompts~\cite{wei-2022:chain} that can enhance the performance of LLMs in the context of few-shot KBQG.

In this work, we propose KQG-CoT framework, which is the first attempt for training-free few-shot KBQG with LLMs. 
As shown in Figure~\ref{fig:kbqg_pipeline}, our framework consists of two main steps, the objects of which are supportive logical forms selection from an unlabeled data pool and  prompt construction.
% To obtain supportive logical forms, we select multiple logical forms as representatives taking account of both syntactic and semantic features.
To acquire coherent logical forms, we employ a clustering technique to carefully choose multiple logical forms that serve as representatives, considering both their syntactic and semantic characteristics.
%These logical forms are further instantiated with a KB so that no ellipsis and ambiguity exists in the logical forms.
To construct prompt, inspired by the principle of CoT~\cite{wei-2022:chain}, we take the selected logical forms as exemplars and write rationales to split the generation of a complete question into multiple steps.
We concatenate the above rationales with the queried logical form to form a prompt, which guides a LLM to outcome a reasoning process of generating a complex question aligning with the logical form. 
We further improve KQG-CoT to KQG-CoT+ via sorting the supportive logical forms by complexity.

As previous methods rely heavily on the training instances to fine-tune a KBQG model. 
KQG-CoT does not need numerous logical form question pairs to train the models.
We test the performance of our prompting methods under few-shot setting on three public datasets, namely WebQuestions~\cite{Kumar-ISWC-2019:MHQG}, PathQuestions~\cite{zhou-etal-2018-interpretable}, and GrailQA~\cite{Gu-2020:GrailQA}.
We conduct a comprehensive comparison with a range of commonly used CoT baseline methods including Auto-CoT~\cite{zhang-2023:automatic}, Active-CoT~\cite{diao-2023:active}, Random-CoT~\cite{Brown-NEURIPS2020:icl} and so on.
% \leicomment{Could we tell readers the reason why we choose the above baselines rather than least-to-most, decomposition prompting, and successive prompting?}
The experimental results show that we can outperform all of them with an observable margin.
Besides, we also compare with a set of SoTA systems trained with full data or few data.
Our few-shot method could achieve competitive results to the full training methods.
Remarkably, our few-shot method could surpass existing few-shot SoTA results of PathQuestions dataset by $18.25$, $10.72$ and $10.18$ absolute points on BLEU-4, METEOR and ROUGE-L, respectively. 
%\leicomment{These parts show our paper tends to discuss on zero-shot instead of few-shot. However, related work section mainly talk about few-shot-based works.}

KQG-CoT provides a simple but effective solution to few-shot KBQG problem, we expect it could serve as an important baseline for future investigation to KBQG tasks under low-resource scenarios.

%We conducted experiments on three commonly used knowledge base question answering (KBQA) datasets, namely WebQuestions (WQ)~\cite{Kumar-ISWC-2019:MHQG}, PathQuestions (PQ)~\cite{zhou-etal-2018-interpretable}, and GrailQA (GQ)~\cite{Gu-2020:GrailQA}. The results of the experiments demonstrate that our method significantly outperforms the baseline models, and achieves performance levels that are comparable to those of models that require extensive training data.

%  六、罗列本篇论文的贡献

Our main contributions are summarized as follows:

\begin{itemize}
    \item By encoding and clustering the skeletons of logical forms, we successfully retrieved supportive logical forms that are particularly suitable for constructing effective prompts.
    \item We reorganized the sequence of examples and utilized the CoT method to construct prompts that are highly effective for large language models.
    \item The experimental results indicate that our method surpasses the baseline by a significant margin and achieves performance levels that are comparable to fine-tuned methods.
\end{itemize}

\section{Related Work}

% \yscomment{Please summarize related work under this topic}

\noindent \textbf{Knowledge Base Question Generation.}
The early approaches for KBQG tasks are template-based methods.
\citeauthor{berant-etal-2013:semantic}~(\citeyear{berant-etal-2013:semantic} and \citeauthor{talmor-berant-2018:web}~(\citeyear{talmor-berant-2018:web}) utilized search engines and manual annotation to construct the natural language questions based on logical forms.
%While template-based methods are simple to implement, they rely on manual intervention, which is hard to be scaled up.
However, template-based methods rely on manual intervention, which is hard to be scaled up.
With the advancement of deep neural networks, neural network-based methods have emerged as a prominent and widely adopted approach.
\citeauthor{Kumar-ISWC-2019:MHQG}~(\citeyear{Kumar-ISWC-2019:MHQG}) and \citeauthor{Chen-2020:TowardSG}~(\citeyear{Chen-2020:TowardSG}) proposed end-to-end models based on Transformer and Graph2seq models, which are capable of generating complex, multi-hop questions based on a subgraph.
Follow-up studies~\cite{fei-etal-2022:lfkqg,guo-etal-2022:dsm} developed more complicated models for KBQG, which ensure the relevance between the generated questions and subgraphs.
%\citeauthor{fei-etal-2022:lfkqg}~(\citeyear{fei-etal-2022:lfkqg}) developed a controlled generation framework for KBQG, ensuring the relevance between generated questions and subgraphs.
%\citeauthor{guo-etal-2022:dsm}~(\citeyear{guo-etal-2022:dsm}) employed graph contrastive learning to retrieve semantically similar subgraphs and combines it with a meta-learner to generate questions.
\citeauthor{xiong-2022:AutoQGS}~(\citeyear{xiong-2022:AutoQGS}) proposed a method for low-resource KBQG, where an auto-prompter is developed to paraphrase a logical form into a description, so that a pre-trained language model can be fine-tuned with the augmented data.
% ~\yscomment{Please check my description if it is accurate.}. 
Our work is different from this one as our method focuses on solving few-shot KBQG challenge with frozen LLMs.
\noindent \textbf{Few-shot Learning for Text Generation.} In recent years, significant progress has been made in the field of few-shot learning for text generation. 
%Researchers have been actively investigating various techniques and models to address the challenge of generating meaningful and coherent text with limited training data.~\cite{Wang-2020:Generalizing}.
One line of work develops meta-learning frameworks for text generation~\cite{Mi-2019:Meta-Learning,madotto-etal-2019:personalizing,Zeng-2021:Domain,Hospedales-2022:Meta-Learning}, which aims to acquire an optimal initialization that enables accurate and rapid adaptation to a new task, even when limited data is available.
%~\cite{Li_Wang-2020:MetaMT,lakumarapu-etal-2020:end,Park-2021:inproceedings} facilitate the learning process, one strategy is to train a meta-learner that optimizes the optimizer of the original network in order to update parameters more effectively.
Other line of work proposes different augmentation algorithms to synthesize the data for training~\cite{SONG-2020:Learning,zhao-etal-2022:improving}, so that conventional text generation models can be applied to the augmented data.
Most recently, LLMs are leveraged to solve few-shot text generation tasks such as text summarization~\cite{yang-2023:exploring,zhang-2023：benchmarking,liu-2023:gpteval}, machine translation~\cite{wang-2023:documentlevel,hendy-2023:good}, dialogue generation~\cite{zhang-2023:prompting,valvoda-etal-2022:prompting,kang-2022:knowledgeconsistent} and so on.
%it has been demonstrated that LLMs are capable of effectively tackling few-shot text generation tasks~\cite{brown-2020:language}.
% ~\cite{Ma-2022-aaai:Switch-GPT} % add more paper later
%Our work is also a text generation task based on large pre-trained language models, but what sets it apart is the incorporation of a knowledge graph.
%Our work employs LLMs in few-shot KBQG tasks, where the input is in the format of a program structure.
There is no existing study applying LLMs to few-shot KBQG tasks.

\noindent \textbf{In-Context Learning with LLMs.}
Without gradient updates, In-Context Learning (ICL) effectively tackles a wide range of NLP tasks by incorporating a small number of prompted examples as part of the input~\cite{ruis-2023:large} to help LLMs understand the tasks.
Multiple studies~\cite{su-2022:Selective,rubin-etal-2022:learning} explored the selection of examples that are similar to the query during prompt construction.
Recent researches~\cite{lu-etal-2022:fantastically,liu-etal-2022:makes,diao-2023:active,wang-2023:selfconsistency} highlight that the order of these examples in the prompt has a substantial influence.
CoT is a prompting strategy decomposing complex tasks into sub-tasks, helping the model to derive the correct answers progressively~\cite{wei-2022:chain,zhou-2023:leasttomost}.
It has been widely used in mathematical word problem solving, common-sense reasoning, and symbolic reasoning.
Our work incorporates CoT strategy into KBQG tasks, where iterative process enables LLMs to ultimately obtain a complex question aligning with the logical form.

\section{Methodology}

\subsection{Problem Formulation}

A KB consists of a set of triples.
A logical form is a structural expression of a subgraph in the KB, which may consist of complex operations (e.g., aggregation, comparative and superlative) and can be utilized to execute against a KB.
% KBQG is a task given a logical form and a system needs to generate a natural language question based on KBs that are semantically consistent with the logical form.
The task of KBQG requires a system to generate a natural language question when given a logical form and the corresponding KBs with consistent semantics.

% KBQG is a challenging task where a system is provided with a logical form and tasked with generating a coherent and meaningful natural language question from KBs that aligns semantically with the given logical form.

%The data used for the few-shot KBQG task consists of a small subset of sub-expressions from the knowledge base, along with corresponding questions. On a standard few-shot CoT prompting setting~\cite{wei-2022:chain} for LLMs, which includes K in-context examples$ \{x_i,r_i,y_i\}_{i=1}^K$ and and an $x_{test}$ as input. The output $y_{test}$  and \textit{Prompt} are as follows:
% \begin{equation} \label{eq:genarate}
%  \arg \max  G\left(r_{\text {test }},y_{\text {test }} \mid Prompt \oplus x_{\text {test }}\right)
% \end{equation}  
% \begin{equation}\label{eq:prompt}
%  Prompt = [(x_3,r_3,y_3) \oplus (x_K-1,r_K-1,y_K-1) \cdots ]  
% \end{equation} 
%where $r$ is the CoT reason step and $\oplus$ represent the concatenation operation.Equation ~(\ref{eq:prompt}) contains K examples in a special order to build prompt.

\subsection{Method Overview}

\begin{figure*}[ht!]
\centering
\includegraphics[width=1.0\textwidth]{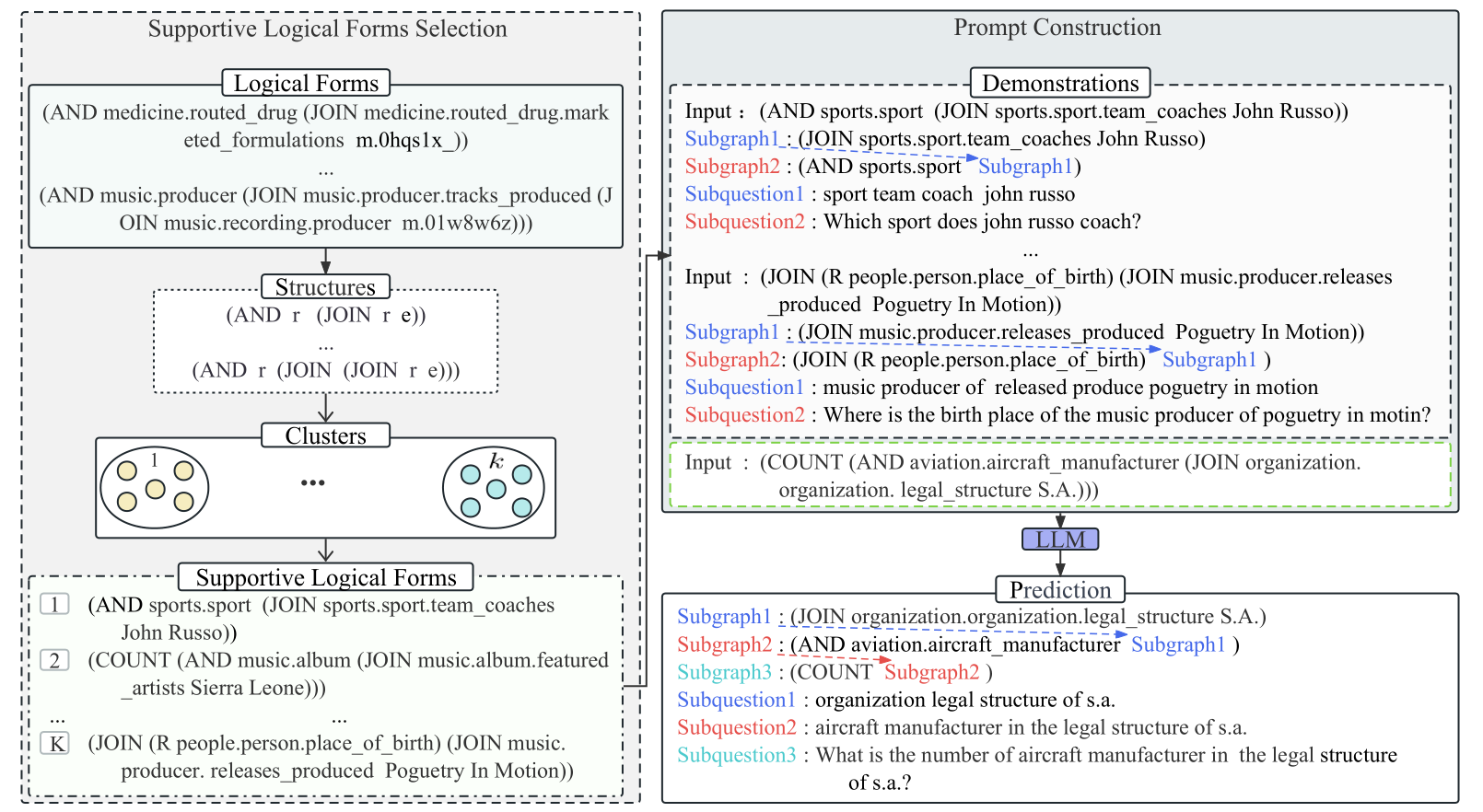}
%\vspace{-1.5cm}
\caption{KQG-CoT framework. The supportive logical forms are selected from an unlabeled data pool by extracting the structures, clustering the structures and sampling the most representative ones. 
%Then the supportive logical forms are instantiated by filling the surface name of the entities.
A total of $k$ demonstrations are automatically constructed using reasoning chains. 
The tested logical form is appended to the demonstrations to form the complete prompt, which can elicit the LLM to generate a series of subquestions sequentially from simple to complicated.
% would be fed into a LLM for generating a question.
Finally, the last subquestion can be extracted as the final prediction.
% \yscomment{replace ``retrieving supportive demonsrations'' to ``selecting supportive logical forms'', ``supportive demonstrations '' to ``supportive logical forms'', ``constructing chain prompt'' to ``constructing prompt'', ``prompt'' to ``demonstrations''}
% %\leicomment{This figure is important. We should improve figure presentation. At least, I thought the retrieving xxx part doesn't show retrieving. It mainly show how to construct a supp demo. What's the meaning of different colors on the constructing part. }
% \yscomment{If $k$ in text is lowercase, please use $k$ in the figure instead.}
} 
\label{fig:framework}
\end{figure*}

Recently, the LLM has shown its impressive in-context few-shot learning capabilities.
Instead of fine-tuning a pre-trained model to adapt it to a downstream task,
we can simply apply it to a new task with a few examples as prompt during inference~\cite{yang:aaai2022,li:acl2023}.
% For KBQG task, we need to \textbf{select supportive logical forms} from unlabeled data.
% % \leicomment{Where is the term supportive demos from?}
% These supportive examples should be representative to cover the possible syntax of logical forms.
% To this end, we  encode the structure of the logical forms and perform clustering as well as sampling to select top-$k$ supportive logical forms.
% After the examples are identified, we leverage LLMs to generate natural language questions.
% To tackle logical forms with complex structure, Chain-of-Thought prompting~\cite{wei-2022:chain} is conducted to explicit the reasoning step.
% We \textbf{construct prompt} to produce sub-questions as rationales which eventually results in a complicated question.
%  % \leicomment{What's the difference between cot prompting and chain prompts} 
For the KBQG task, we adopt a two-stage method to design CoT prompts, which effectively enable the LLM to comprehend complex logical forms and generate questions.
Concretely, the first stage \textbf{Supportive Logical Forms Selection} focuses on identifying supportive examples that represent various syntax patterns of logical forms. To accomplish this, we encode the structure of logical forms, perform clustering, and employ sampling techniques to select top-$k$ supportive logical forms.
Once these supportive examples are selected, we leverage LLMs with CoT prompts to generate natural language questions. This leads us to the second stage, \textbf{Prompt Construction}, which involves producing sub-questions as rationales. Through this process, we can ultimately formulate a complex question that adequately captures the semantic of the logical form.
A schematic diagram of our method is displayed in Figure~\ref{fig:framework}.

\subsection{Supportive Logical Forms Selection}
\label{sec:retrieve_lf}

\citeauthor{zhang-2023:automatic}~(\citeyear{zhang-2023:automatic}) has shown that when constructing demonstrations, we need to mitigate the effect of few-shot CoT errors by differentiating the design of demonstrations.
In KBQG tasks, supportive logical forms are those that can cover diverse logical rules, so as to offer more syntax information for LLMs to generate questions.
% However, it is not trivial to adapt the existing CoT methods because unlike the narrative inputs which can directly be encoded by textual pre-trained models. 
% The logical form is a combination of program structures and schema items (i.e., entities and relations). 
% Hence, we need to take both aspects into consideration when we encode the logical forms.
Unlike the narrative inputs, the logical form is a combination of program structures and schema items (i.e., entities and relations). Therefore, it is essential to take both aspects into consideration when selecting supportive logical forms. In our approach, we utilize \textbf{Structure Encoding and Clustering}, followed by a \textbf{Logical Form Sampling} process to select supportive logical forms.

%In this stage, we need to retrieve diversity support logical forms that are relevant to the Prompt we want to generate. These logical forms can come from different domains, resources, or contexts. The purpose of these logical forms is to provide diverse perspectives, so as to offer more inspiration and possibilities when constructing the Prompt. As the few-shot CoT reasoning is sensitive to both the diversity and order of the samples, and finding the best prompt is an NP-hard problem, we have opted to construct optimal prompts in two stages. The first stage is Skeleton Encoding and Clustering, while the second stage is Logical Form Sampling and Reordering.

\noindent \textbf{Structure Encoding and Clustering}.
To ensure the logical forms can be drafted for unseen questions, we extract their structures by converting the schema items into symbolic variables.
Specifically, we keep the grammars in the logical form unchangeable.
Then, we replace the relation with symbol ``\textit{r}'' and we replace the entity with ``\textit{e}''.
This structure is also known as a abstract query graph~\cite{chen:arxiv2021}, which reflects the topology and the component classes of logical forms.
For instance, the raw logical form is:

\begin{itemize}[label={}, labelsep=0pt, leftmargin=10pt]
\item {\small (AND \texttt{medicine.routed\_drug} \\
(JOIN \texttt{medicine.routed\_drug.marketed\_formulations}  m.0hqs1x))}.
\end{itemize}
It becomes the following structure after conversion:
\begin{itemize}[label={}, labelsep=0pt, leftmargin=10pt]
\item {\small (AND \texttt{r} (JOIN \texttt{r}  e))}.
\end{itemize}

Once we have obtained the structure of the logical forms, which filters out the semantic meaning of the logical forms.
We encode the structure representation into a fix-length embedding.
In detail, we view the structure as a sequence of tokens.
%We compute a embedding for each sequence by sentence-transformers~\footnote{https://huggingface.co/sentence-transformers/all-MiniLM-L6-v2}.
We encode the contexts of the sequence with Sentence-Transformers~\cite{reimers-gurevych-2019-sentence}, which is an advanced model for text embedding.
%Sentence-Transformers use deep neural networks to model the context of sentences. 
%This means that the model can understand the relationships between words in a sentence, including word order, dependencies, and more.
%in order to better capture the meaning of the sentence. 
The encoded vectors are well-suited for calculating the similarity between sentences.
% ~\yscomment{Is there any reason that we can say about its benefits to encoding the grammar?}
% \lyyresponse{Can we add the following paragraph? "According the syntax rules of logical forms, it becomes apparent that the main structural patterns of logical forms closely correspond to the primary structure of a sentence. Additionally, the remaining relational and topical entities can be transformed into sentence modifiers and subjects. Therefore, encoding the structure of logical forms enables us to efficiently and swiftly filter out the logical forms that correspond to different types of sentences."}
%~\yscomment{I actually mean why we choose to use Sentence Transformer to encode the structure, is there any thing we can say about its benefits to the grammar encoding?}.
We extract the final hidden state of as the vectorized representation of the sentence.
%By considering the final hidden state of the last input as the semantic vector of the sentence, we obtain a 384-dimensional sentence vector.
%...\yscomment{Please describe how you embed the structure in detail. What pre-trained models you use to encode the sequence? Which vector you use as the structure representation?}
After that, we utilize the K-means~\cite{hartigan-1979:algorithm} clustering algorithm to group the encoded structure into $k$ clusters based on their syntactic similarity.

\noindent \textbf{Logical Form Sampling}.
Each cluster contains a group of logical forms with the similar structure, we randomly pick up a structure from each group and obtain $k$ representative structures.
As each structure may correspond to multiple logical forms.
We further identify $k$ logical forms with distinct semantics deriving from the $k$ selected structures.
To this end, we iteratively sample logical forms holding the maximum diversity of semantics.
% For each structure, there are multiple logical form associated with it.
Specifically, for the first logical form, we randomly pick up one from the candidates.
Then we search logical forms for another structure. 
We greedily pick up a candidate with least semantic similarity to the selected logical forms, where the similarity is measured by the encoding of the original logical forms.
We repeat the process until we have gone through $k$ structures as shown in Figure~\ref{fig:framework}.

To help the LLMs fully understand the logical forms, we substitute the entities in the original logical forms with their surface names in the KB. In this way, we obtain $k$ supportive logical forms.

\subsection{Prompt Construction}

Since some logical forms have complicated semantics and even nested syntactic structures are included. 
Following the CoT method, we construct a reasoning chain prompt based on the supportive logical forms retrieved above.
For each example, we need to generate a reasoning chain based on logical forms to elicit LLMs generate questions from simple to complicated.
To this end, we hold two criteria when constructing reasoning chains: 

\begin{enumerate}[label=(\roman*)]
\item The templates should break up the generation of a complicated question into a step by step process. 
\item The templates should clearly identify the sub-component in a logical form that requires LLMs to focus on for each step.
\end{enumerate}

Therefore, we first break down a logical form in a nested manner, where the follow-up logical forms include the preceding logical forms.
Specifically, the first step usually generates a simple question querying one-hop relation from the topic entity.
The second step usually generates a question querying two-hop relation chain involving the above one-hop relation.
As we can see from Figure~\ref{fig:framework}, the first step of prompt parses the entire logical form into one-hop relation \textit{subgraph1} ``(AND \texttt{sports.sport.team\_coaches John Russo})'' which leads to a simple \textit{subquestion1} ``\textit{sport team coach john russo }''.
%\yscomment{I feel this is not a grammatically correct sentence.}\lyyresponse{The intermediate subquestion should be approximately a phrase, as long as it contains sufficient information without generating any ambiguity. Strict grammatical accuracy is not required, as LLM will make adjustments to ensure the final subquestion is grammatically correct.}.
The second step includes the parsed logical form appended to the previous step as a component and generates question ``\textit{Which sport does john russo coach?}'' based on the \textit{subgraph2} and \textit{subquestion1}.
As a result, we continuously expand the logical form until a complete question is formed. 
This step-by-step  process ensures that the generated question is semantically coherent and grammatically accurate.

During inference, we concatenate all the demonstrations and queried logical form as the final prompt.
Based on the example in Figure~\ref{fig:framework}, the prompt includes ``\textit{Input: (AND ... Input: (JOIN ... Input: (COUNT ... S.A.}''. 
After receiving the prompt, LLMs outcome the predictions that clarifies the intermediate generation steps of \textit{subquestion1}, \textit{subquestion2}, and \textit{subquestion3}.
And the last subquestion will be our final predicted question, which is ``\textit{What is the number of aircraft manufacturer in the legal structure of s.a. ?}''.

\section{Experiment}
\label{sec:expr}

\begin{table}[]
\centering  
\begin{tabular}{ccccc}
\toprule
Dataset & \#Q & \#R & \#E & \#T \\ 
\midrule
WQ & 22,989 & 672   & 25,703 & 2/99/5.8 \\
PQ & 9,731  & 378   & 7,250  & 2/3/2.7  \\
GQ & 64,331 & 3,720 & 32,585 & 1/4/1.4  \\ 
\bottomrule
\end{tabular}
\caption{\label{dataset-static}
Statistics of the evaluated datasets.
\#Q denotes the number of questions. 
\#R and \#E denote the total number of relations and entities, respectively. 
\#T denotes the minimum/maximum/average number of triplets involved in each question.
}
\end{table}

\begin{table*}[]
\centering  
\small
\begin{tabular}{l | c c c c c c c c c}
\toprule
\multirow{2}{*}{Method} &
  \multicolumn{3}{c}{WQ} &
  \multicolumn{3}{c}{PQ} &
  \multicolumn{3}{c}{GQ} \\
 &
  B &
  M &
  R &
  B &
  M &
  R &
  B &
  M &
  R \\ 
  \midrule
Standard Prompt & 
    24.86  & 29.01  & 52.74  & 55.87  & 42.24  & 76.83  & 29.17  & 33.52  & 42.95\\
Random-CoT&    
    25.02 & 29.37 & 53.16 & 56.42 & 42.61 & 77.03 & 29.81 & 33.75 & 43.31  \\
Manual-CoT &
    28.44 & 30.24 & 54.30 & 60.37 & 42.88 & \underline{77.48}& 30.18 & 33.61 & 44.89\\
Active-CoT & 
    26.02 & 29.55 & 54.01 & 58.78 & \underline{43.86} & 76.78 & 30.27 & 33.71 & 44.07\\
Auto-CoT &  
    28.42 & 29.65 & 53.47 & 59.59 & 43.16 & 77.13 & 30.17 &  34.22 & 44.47\\
\midrule
KQG-CoT (Ours) & 
   \underline{28.89} & \underline{30.41} & \underline{54.38} & \underline{60.81} & 
    43.54 &  77.35 & \underline{30.51} & \underline{34.26} &
    \underline{44.91}\\
KQG-CoT+ (Ours) &
    \textbf{29.73} & \textbf{31.08} & \textbf{55.14} & \textbf{61.71} & 
    \textbf{44.27} & \textbf{ 78.41} & \textbf{31.24} & \textbf{34.94} &
    \textbf{45.36}\\ 
\bottomrule
\end{tabular}
\caption{
 Few-shot evaluation of existing prompting methods with Frozen LLMs on three KBQG datasets.
 The best and second best results are boldfaced and underlined respectively.
%  \yscomment{2. Can you fine-tune the methods as I notice WQ-R, PQ-M, GQ-M are really close to the second best results.}
}
\label{tab:main}
\end{table*}

In this section, we first introduce the KBQG datasets used to evaluate the performance of our proposed method and the comparable baseline methods.
Next, we present the implementation details and demonstrate the experimental results.

\subsection{Data and Metrics}

We evaluate our prompting method on the following three public datasets:

\noindent \textbf{WebQuestions (WQ)}~\cite{Kumar-ISWC-2019:MHQG}\footnote{\url{https://github.com/liyuanfang/mhqg}} is a KBQG dataset combining instances from WebQuestionsSP~\cite{serban-etal-2016-generating} and ComplexWebQuestions~\cite{talmor-berant-2018-web}.
It provides questions, answers, and annotated subgraphs.
This dataset is commonly evaluated in existing work~\cite{guo-etal-2022:dsm}.

\noindent \textbf{PathQuestions (PQ)}~\cite{zhou-etal-2018-interpretable}\footnote{\url{https://github.com/zmtkeke/IRN}} is a commonly used KBQG dataset constructed from a KBQA dataset.
It contains questions inquiring a chain of relations, wherein the path between the topic entities and answer entities is $2$-hop or $3$-hop.

\noindent \textbf{GrailQA (GQ)}~\cite{Gu-2020:GrailQA}\footnote{\url{https://dki-lab.github.io/GrailQA/}} is a large-scale KBQA dataset built on Freebase, which covers $86$ domains.
It covers complex questions which require counting, ranking and even superlative inquiry.
Each question is associated with a s-expression, which can be viewed as a logic form.

We collect the annotated the logic form from the training set as the data pool and leave the original questions untouched.
The questions in the validation or test set are sampled to evaluate our method.
Statistics of evaluated datasets are shown in Table~\ref{dataset-static}.
%\yscomment{I am afraid that people may argue our comparison with other systems are not fair.}

Following previous KBQG studies, we rely on a set of well-established metrics as for KBQG evaluation: BLEU-4~\cite{papineni-etal-2002-bleu}, METEOR~\cite{banerjee-lavie-2005-meteor} and ROUGE-L~\cite{lin-2004-rouge}. 
BLEU-4 and ROUGE-L can be viewed as precision and recall for text generation tasks, respectively.
METEOR is a comprehensive metric beyond exact matches, which also accounts for partial matches and variations in word order.
We denote them as \textbf{B}, \textbf{M} and \textbf{R}, respectively.
%making it a robust evaluation measure for translation performance. \yscomment{Please complete the sentence.}
%~\lyyresponse{Written.}
% that considers multiple aspects of text quality, such as precision, recall, and alignment between the generated output and reference translation. It goes

%To demonstrate the effectiveness of our method, we conducted experiments on three widely used datasets for KBQA tasks: WebQuestions (WQ)(\citealp{Kumar-ISWC-2019:MHQG}), PathQuestions(PQ) (\citealp{zhou-etal-2018-interpretable}), and GrailQA(GQ) (\citealp{Gu-2020:GrailQA}).All three datasets utilize Freebase as their underlying knowledge base.

%The WebQuestions and PathQuestions datasets provide questions, answers, and relevant triples (subject, relation, object) extracted from the Freebase knowledge base to provide contextual information for the questions, aiding the algorithms in understanding and answering the questions.The number of instances in the training, validation, and test sets for both datasets are as follows: 18,989/2,000/2,000 for WebQuestions, and 9,793/1,000/1,000 for PathQuestions.

%GrailQA covering 32,585 entities,3,720 relations across 86 domains. It is designed to test three levels of generalization of KBQA models: I.I.D., compositional, and zero-shot.

\subsection{Comparable Methods}

We denote our prompting method as \textbf{KQG-CoT}.
Previous studies~\cite{lu-etal-2022-fantastically} have proven that the order of the exemplars is significant to the prompt results, we implement an improved version by sorting the demonstrations from short to long after sampling.
We denote this method as \textbf{KQG-CoT+}.

As there is no existing attempt for few-shot KBQG tasks with LLMs, we adopt five general prompting methods under few-shot scenarios as our baselines.

\noindent \textbf{Standard Prompt}~\cite{Brown-NEURIPS2020:icl} 
is a standard prompting method of in-context learning, where $k$ random logical forms and questions are concatenated to form the prompt.
The prediction is one-step generation.
%is a standard prompting method of in-context learning involves randomly selecting K logical forms and generating their corresponding questions. 
%These K logical forms and questions are then concatenated together to form the prompt.

\noindent \textbf{Random-CoT} is an intuitive CoT prompting baseline where $k$ logical forms are randomly selected from the data pool and we follow the original work~\cite{Brown-NEURIPS2020:icl} to describe the sub-task in a narrative.

% is a straightforward method for constructing prompts is to randomly select K logical forms and then employ the Chain-of-Thought (CoT) approach, similar to our method, to generate questions based on them.

\noindent \textbf{Manual-CoT}~\cite{wei-2022:chain} is a CoT prompting with $k$ human-written exemplars as demonstrations and the sub-task is presented in narratives.
% is a CoT prompting involves providing K human-written exemplars that consist of a series of intermediate reasoning steps.

%\textbf{Active-CoT}~\cite{diao-2023:active}: Strictly speaking, this is not a method that adheres to the standard few-shot setup. It involves randomly selecting some logical and question pairs to construct a Cot prompt. The validation set is created by extracting one logical and question pair from each skeleton. The ( 1 - BLEU-4 score )  is used as the uncertainty value for each skeleton. Multiple experiments are conducted to select the top-K logical forms based on the average uncertainty value.

\noindent \textbf{Active-CoT}~\cite{diao-2023:active} is an ensemble framework for CoT prompting.
The multiple logical forms are randomly selected as a validation set.
Then multiple measurements (e.g., disagreement, variance) are leveraged as the uncertainty value for each logical form to produce the final question.

\noindent \textbf{Auto-CoT}~\cite{zhang-2023:automatic} automatically constructs prompt by selecting $k$ demonstrations with a cluster-based algorithm and the sub-task is presented in narratives.
We simply adopt the prompting method to KBQG tasks by encoding all logical form in a textual way.

%is an automatic exemplar construction method is utilized, which involves applying clustering techniques to sample logical forms. This method then generates chains of intermediate subquestions based on the subgraph.

\subsection{Implementation Details}
For encoding of logical forms, we utilize all-MiniLM-L6-v2\footnote{\url{ https://huggingface.co/sentence-transformers/all-MiniLM-L6-v2}} checkpoint from the Sentence-Transformers library in Huggingface for effective encoding.
As this is a few-shot scenario, we manually write the rationales for the $k$ demonstrations in the chain prompt.
We utilize \texttt{text-davinci-003} from OpenAI API\footnote{\url{ https://openai.com/blog/openai-codex/}} to generate questions and set the number of clusters as $k=12$\footnote{Detailed prompt design of KQG-CoT+ is presented in Appendix~\ref{instruction_prompt}.}.

\subsection{Main Results}
% \usepackage{multirow}

\begin{table}[ht!]
\centering  
\small
\resizebox{\linewidth}{!}{
\begin{tabular}{l | c c c }
\toprule
\multirow{2}{*}{Method} &
  \multicolumn{3}{c}{WQ}  \\
 & B & M & R \\ 
  \midrule 
\multicolumn{4}{c}{\textit{Full Training}} \\
L2A~\cite{du-etal-2017-learning} & 6.01 & 26.95& 25.24 \\
Transformer~\cite{Vaswani-NIPS2017:attention} & 8.94 &13.79 &32.63 \\
MHQG~\cite{Kumar-ISWC-2019:MHQG} & 11.57 & 29.69 & 35.53 \\
BiGraph2Seq~\cite{Chen-2020:TowardSG} & 29.45 & 30.96 & 55.45 \\
T5-Large~\cite{raffel-2020:exploring} & 28.78 & 30.55 & 55.12\\
JointGT~\cite{ke-etal-2021:jointgt} & 30.02 & 32.05 & 55.60 \\
IGND~\cite{fei-2021:iterative} & 30.62 & 31.41 & 55.82 \\
LFKQG~\cite{fei-etal-2022:lfkqg} & \textbf{31.66} & \textbf{32.69} & 56.75 \\
DSM~\cite{guo-etal-2022:dsm} & 28.62 & - & \textbf{64.25} \\
\midrule
\multicolumn{4}{c}{\textit{Few-shot Evaluation}} \\
KQG-CoT & 28.89 & 30.41 & 54.87\\ 
KQG-CoT+ & 29.73 &31.08 &55.46 \\ 
\bottomrule
\end{tabular}
}
\caption{
 Comparison between few-shot evaluation of KQG-CoT/KQG-CoT+ and full-trained evaluation of other systems on WQ.
 %\yscomment{Do we have more baselines the results of which are lower than us?}
}
\label{tab:wq-result}
\end{table}

\begin{table}[ht!]
\centering  
\small
\resizebox{\linewidth}{!}{
\begin{tabular}{l | c c c }
\toprule
\multirow{2}{*}{Method} &
  \multicolumn{3}{c}{PQ}  \\
  &
  B &
  M &
  R \\ 
  \midrule
\multicolumn{4}{c}{\textit{Full Training}} \\
L2A~\cite{du-etal-2017-learning} & 17.00 & 50.38 & 19.72 \\
Transformer~\cite{Vaswani-NIPS2017:attention} & 56.43& 43.45 &73.64 \\
MHQG~\cite{Kumar-ISWC-2019:MHQG} & 25.99 & 33.16 & 58.94 \\
BiGraph2Seq~\cite{Chen-2020:TowardSG} & 61.48 & 44.57 & 77.72 \\
AutoQGS~\cite{xiong-2022:AutoQGS} & 65.13 & 47.50 & 76.80 \\
T5-Large~\cite{raffel-2020:exploring} & 58.95 & 44.72 & 76.58 \\
IGND~\cite{fei-2021:iterative} & 61.69 & 45.11 & 77.28 \\
LFKQG~\cite{fei-etal-2022:lfkqg} & 63.92 & 46.91 & 78.40 \\
JointGT~\cite{ke-etal-2021:jointgt} &  \textbf{65.89} & \textbf{48.25} & 78.87\\
DSM~\cite{guo-etal-2022:dsm} & 61.03 & - & \textbf{86.06} \\ \midrule
\multicolumn{4}{c}{\textit{Few-shot Evaluation}} \\
BiGraph2Seq~\cite{Chen-2020:TowardSG} & 1.01 & 4.99 & 12.07\\
JointGT~\cite{ke-etal-2021:jointgt} & 43.15 & 35.91 & 69.57\\
AutoQGS~\cite{xiong-2022:AutoQGS} & 43.46 & 33.55 & 68.23\\
%\multicolumn{4}{c}{\textit{Few-shot Evaluation}} \\
KQG-CoT & 60.81 & 43.54 & 77.35\\ 
KQG-CoT+ & 61.71  & 44.27  & 78.41\\ 
\bottomrule
\end{tabular}
}
\caption{
Comparison between few-shot evaluation of KQG-CoT/KQG-CoT+ and few-shot/full-trained evaluation of other systems on PQ.
}
\label{tab:pq-result}
\end{table}

\begin{table*}[t!]
\small
\centering
\begin{tabular}{lll}
\hline
\hline
\multicolumn{3}{l}{\begin{tabular}[c]{@{}l@{}}\textbf{Input}: (AND military.military\_conflict (JOIN military.military\_conflict.force\_strengths (JOIN \\      \qquad \quad     (R military.military\_resource.conflicts) Bendix AN/FPS-20)))\end{tabular}} \\
\multicolumn{3}{l}{\textbf{Manual-CoT}: Which military conflict involves the bendix an/fps-20 and what are its force strengths?}  \\
\multicolumn{3}{l}{\textbf{Active-CoT}: What military conflict has force strengths using bendix an/fps-20?}                       \\
\multicolumn{3}{l}{\textbf{Auto-CoT}: What are the force strengths in the bendix an/fps-20 military conflict?}                    \\
\multicolumn{3}{l}{\textbf{KQG-COT+}: Which military conflict has force strengths with bendix an/fps-20?}                \\
\multicolumn{3}{l}{\textbf{Ground Truth}: Which military conflict has force strengths with conflicts bendix an/fps-20?}          \\ \hline
\multicolumn{3}{l}{\begin{tabular}[c]{@{}l@{}}\textbf{Input}:(AND measurement\_unit.measurement\_system (JOIN measurement\_unit.measurement\_system.\\    \qquad \quad   heat\_capacity\_units Joule per kelvin))\end{tabular}} \\
\multicolumn{3}{l}{\textbf{Manual-CoT}: What is the measurement system that uses joules per kelvin for heat capacity units?}       \\
\multicolumn{3}{l}{\textbf{Active-CoT}: What is the measurement system for heat capacity units of joule per kelvin?}             \\
\multicolumn{3}{l}{\textbf{Auto-CoT}: Which measurement system uses joule per kelvin as its heat capacity unit?}                  \\
\multicolumn{3}{l}{\textbf{KQG-COT+}: What measurement system uses joule per kelvin as a units to measure heat capacity?} \\
\multicolumn{3}{l}{\textbf{Ground Truth}: What system uses joule per kelvin as the unit to measure heat capacity?}                \\ \hline
\end{tabular}
\caption{\label{case-study}
Illustrative examples from KQG-CoT+ and baseline methods on GQ. 
}
\end{table*}

\textbf{Comparison with Baselines.}
Table~\ref{tab:main} showcases the experimental results of our methods and baseline approaches.
We have the following observations based on it:

1) Comparing all CoT prompting methods, in the few-shot setting, our KQG-CoT+ prompting consistently outperforms other method across all KBQG datasets by a remarkable margin.
Specifically, KQG-CoT+ improves the performance of the competitive Auto-CoT by $0.72$ to $2.12$ absolute values for all datasets.
Meanwhile, KQG-CoT also outperforms existing CoT prompting methods on BLEU-4 of all the datasets. 

2) Comparing CoT methods with standard prompting, we notice that all the CoT prompting methods outperform the standard prompting method, which indicates that, to generate questions with complex logic and long dependency, splitting the entire generation task into sub-tasks are crucial for maintaining the coherence and accuracy of the questions.

3) Comparing Auto-CoT, KQG-CoT and KQG-CoT+, even though all these methods adapt clustering to select $k$ demonstrations, KQG-CoT and KQG-CoT+ are more effective as we elaborately design encoding algorithm and prompt templates for KBQG tasks, which makes it fit more into the question generation from the logical forms.

\noindent \textbf{Comparison with Other Systems.}
We further compare our prompting methods with other KBQG systems on the WQ and PQ datasets. According to our knowledge, we are the first to work on the KBQG task using the GQ dataset, so there are no existing methods available for comparison.

In Table~\ref{tab:wq-result}, we can see that with $12$ demonstrations, our method can outperform majority of full-trained systems on WQ dataset, where all training data is leveraged to train a model.
KQG-CoT+ prompting method can achieve $29.73\%$, $31.08\%$ and $55.46\%$ for BLEU-4, ROUGE-L and METEOR respectively, which are close to the SoTA results.

In Table~\ref{tab:pq-result}, we can see that for PQ dataset, our method can still achieve better results than most of existing full-trained KBQG models.
Compared with existing methods under few-shot settings, our methods can significantly improve the BLEU-4 over AutoQGS by around $20$ absolute points.
It is worth noting that AutoQGS takes $0.1\%$ training instances for training and we simply leverage $12$ instances for inference, which highlights superiority of our methods.

% \jncomment{I can see the whole comparison on WQ and PQ in Table 3 and 4, respectively. Why omit GQ? The reviewer may ask this question.}

\subsection{More Analysis}

\begin{table}[ht!]
\centering
\small
\begin{tabular}{lccc}
\toprule
Model        & Synt. & Comp. & Relev. \\ 
\midrule
Ground Truth &    4.88       &   4.92         &      4.91     \\ \midrule
Standard Prompt  &   3.67      &    3.76       &       3.99       \\
Random-CoT         &     4.05       &   4.21         &    4.12       \\
Manual-CoT      &       4.60      &       4.54     &    4.72       \\
Active-CoT        &     4.56      &      4.71      &    4.75       \\
Auto-CoT       &   4.38        &   4.77         &    4.55       \\
\midrule
KQG-CoT+         & \textbf{4.63}      &  \textbf{4.80}      &   \textbf{4.78} \\ \bottomrule
\end{tabular}
\caption{\label{huamn-eval-result}
Results of human evaluations on WQ. 
Synt., Comp. and Relev. denote syntactic correctness, complexity and relevance, respectively.}
\end{table}

\begin{table}[ht!]
\centering 
\centering  
\small
\begin{tabular}{l | c c c }
\toprule
\multirow{2}{*}{Method} &
  \multicolumn{3}{c}{GQ} \\
 & B & M & R  \\ 
  \midrule
KQG-CoT+ & 31.24 & 34.94 & 45.36 \\
\midrule 
(a) w/o CoT & 30.11 & 33.58 & 43.88\\
(b) K-means $\rightarrow$ Random & 29.81 &33.75 & 43.31\\
%(c) K-means $\rightarrow$ Uncertainty &30.27 &33.71 &44.07 \\
%(d) K-means $\rightarrow$ Random & 29.81 &33.75 &43.31\\
%(e) K-means $\rightarrow$ Manual & 30.25 &33.66 &44.99   \\
(c) w/o structure encoding & 30.03 & 33.41 & 43.76 \\
\bottomrule
\end{tabular}
\caption{\label{main-ablation-gq-result}
 Ablation study of our KQG-CoT+ method on GQ.
}
\end{table}

%Evaluation metrics may not accurately reflect the quality of the generated questions, regarding . 
\noindent \textbf{Human Evaluation.}
We further conduct human evaluation by randomly sampling $300$ examples from the test set of WQ dataset.
The generated questions are rated on a scale of $1$ to $5$ considering the aspects of syntactic correctness, complexity, and relevance to the given logical forms.
We ask three annotators to score the generated questions with $1$-point being poor and $5$-point being perfect. The score of each question is averaged over all annotators. We present the results in Table~\ref{huamn-eval-result}, where we can observe a similar trend between human and automatic evaluation.
Our approach outperforms all comparable methods, the evaluated scores of which are close to the ground truth.

\noindent \textbf{Ablation Study.} 
We conduct ablation study to assess the effectiveness of components of our model and display the results in Table~\ref{main-ablation-gq-result}.
We first exclude the CoT reasoning chain, and observe a performance drop of the evaluate metrics.
This indicates that CoT plays an important role in generating complicated questions.
Then we remove the K-means algorithm and randomly select supportive logical forms.
The decrease of the results indicates that our clustering algorithm could provide more diverse logical forms as our demonstrations.
We further encode the entire logical forms without extracting their structures.
The results decrease which indicate that the structure is a significant indicator to obtain the clusters\footnote{The ablation study on WQ and PQ is presented in Appendix~\ref{ablation_study_wq_pq}.}.

%We conducted ablation tests to assess the effectiveness of various components in our model. Firstly, we examined the exclusion of the CoT method during prompt construction. Secondly, we replaced the K-means method with alternative approaches, such as similarity, uncertainty, random, and manual methods, in selecting supportive demonstrations. Additionally, we tested a method where we randomly selected skeletons without encoding them. The analysis results presented in Table ~\ref{main-ablation-gq-result} yielded several significant observations.
%The utilization of prompts constructed using the CoT method demonstrated clear benefits in question generation. By incorporating intermediate generation steps, the model was better guided in producing questions that aligned more accurately with the given input. Furthermore, the K-means method employed in our "Retrieving Supportive Demonstrations" section had a substantial impact. Among the different approaches tested, random selection exhibited the largest decline in performance, while employing the similarity method yielded the best results. However, a slight disparity still remains between our method and the similarity-based approach. Moreover, the effectiveness of our skeleton encoding module was evident. When encoding the entire logical form instead of just the skeleton, the performance was notably poorer, only slightly surpassing random selection methods.

\begin{figure}[ht!]
\centering
\includegraphics[width=0.45\textwidth]{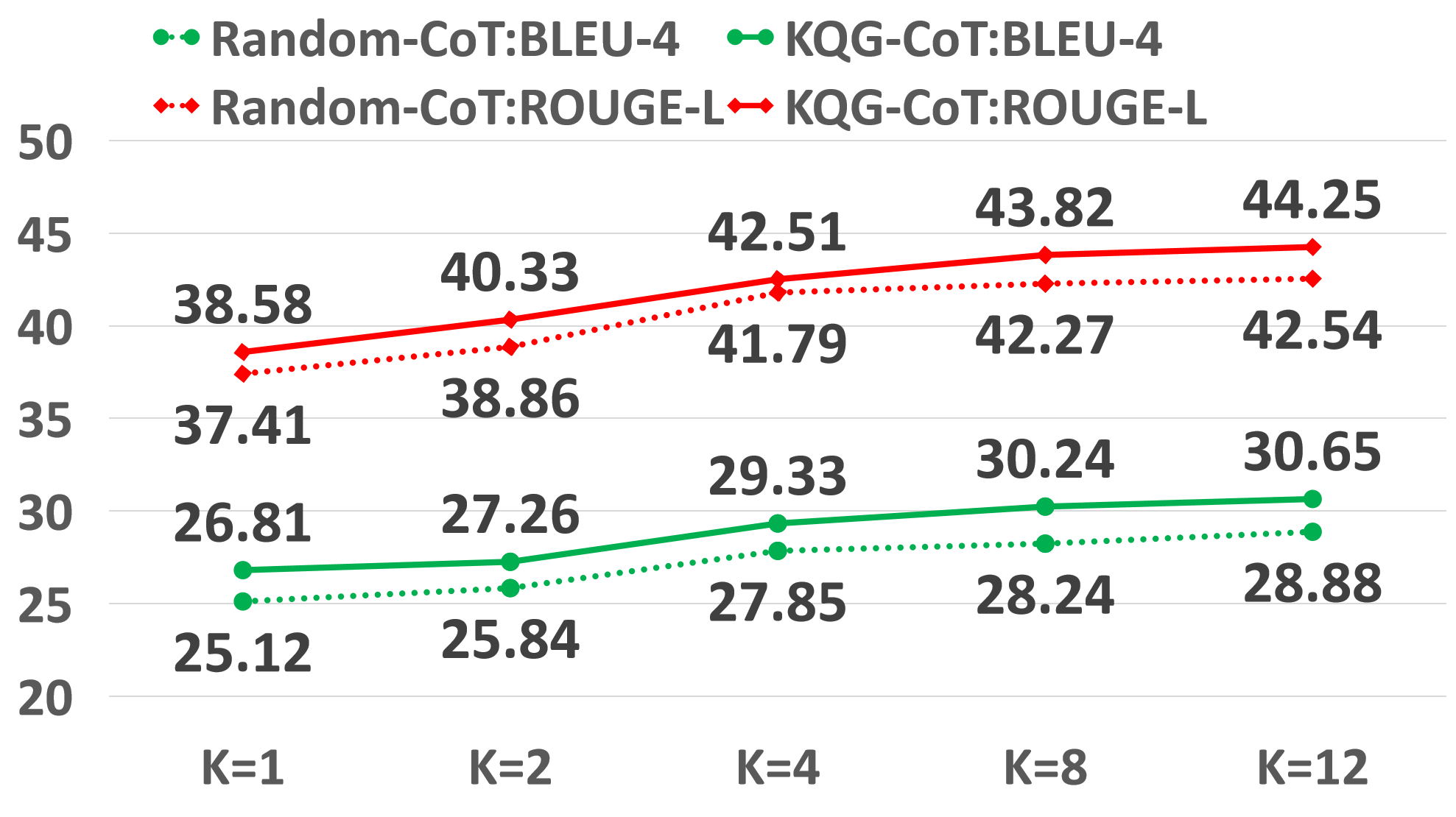}
% \vspace{-0.5cm}
\caption{ The BLEU-4 and ROUGE-L scores of our method and Random-CoT with increasing number of shots on GQ.
%a shot number across 100 tests on the GQ.
% \yscomment{Please add results of baseline.}
}
\label{fig:shot-number}
\end{figure}

\noindent \textbf{Effect of $k$.} 
We investigate the effect of $k$ in Figure~\ref{fig:shot-number}.
As observed, with an increase of the number of demonstrations, both our methods and Random-CoT show increasing BLEU-4 and ROUGE-L scores. 
This indicates that the number of demonstrations is significant in activate the potentials of LLMs.
Compared with Random-CoT, our method shows a larger gain when the value of $k$ becomes large, this indicates that our methods indeed pick up the most representative logical form as the demonstrations.
%However, as K increases, the rate of improvement diminishes, and our methods exhibit a greater advantage over Random-CoT.

\noindent \textbf{Case Study.}
To provide a comprehensive comparison between KQG-CoT+ method and the baseline models on GQ dataset, we present multiple example cases in Table~\ref{case-study}.
Our method elicits the intermediate generation steps and provides more guidance to LLMs so that our KQG-CoT+ generates questions that are grammatically correct and semantically close to the given logical form.
In contrast, baseline methods may encounter issues such as inconsistency in the logical form, misplaced modifiers, or unsmooth expressions.

\noindent \textbf{Effectiveness of Structured Encoding and Clustering.}
To demonstrate the effectiveness of the proposed Structured Encoding and Clustering in selecting diverse structures, we conducted a quantitative assessment of the average semantic similarity between the logical forms extracted using our method and the baseline method at K=8 on the GrailQA dataset. The results are presented in Table ~\ref{effectiveness-of-cluster-result}. The data from the initial segment, shown in the table below, reveals that the logical forms chosen by our method exhibit a lower average semantic similarity. When viewed collectively, these findings offer strong evidence for the efficacy of our proposed approach.

\begin{table}[ht]
\centering
\small
\begin{tabular}{cc}
\hline
Method     & Average\_similarity \\ \hline
Random     & 0.285               \\ \hline
Active-CoT & 0.274               \\ \hline
Auto-CoT   & 0.265               \\ \hline
KQG-CoT    & 0.252               \\ \hline
\end{tabular}
\caption{\label{effectiveness-of-cluster-result}
 The average semantic similarity between the logical forms of different methods.}
\end{table}

\noindent \textbf{Impact of Sorted Order.}
To assess the impact of the sorted order of demonstrations in KQG-CoT+, we compared the performance of Auto-CoT and Active-CoT using the same sorted order of demonstrations in KQG-CoT+ (i.e., Auto-CoT+ and Active-CoT+) and conducted experiments on the GrailQA dataset . The Table ~\ref{role-of-sorted-order-result} shows that, compared to the Active-CoT+ and Auto-CoT+ methods, our proposed KQG-CoT+ method still exhibits significant improvements.

\begin{table}[ht]
\centering
\small
\begin{tabular}{cccc}
\hline
Method      & B     & M     & R     \\ \hline
Active-CoT+ & 30.40 & 34.04 & 44.22 \\ \hline
Auto-CoT+   & 30.52 & 34.59 & 44.77 \\ \hline
KQG-CoT+    & 31.24 & 34.94 & 45.36 \\ \hline
\end{tabular}
\caption{\label{role-of-sorted-order-result}
 The result data for Auto-CoT+, Active-CoT+, and KQG-CoT+ on the GrailQA dataset. }
\end{table}

\noindent \textbf{ KQG-CoT Improve KBQA Task.}
To confirm the efficacy of our approach in enhancing the performance of KBQA methods, we initiated a data augmentation procedure for the WebQuestions dataset. It's important to highlight that the augmented dataset was merely half the size of the original dataset. Next, we trained the KBQA method RnG-KBQA~\cite{ye-etal-2022-rng} by combining the augmented and original datasets, resulting in the improved version called RnG-KBQA+. The results, as outlined in Table~\ref{kbqg-improve-kbqa-result}, demonstrate that we conducted a relatively straightforward augmentation on a limited dataset subset. Nevertheless, the F1 score of the original KBQA method witnessed a notable increase of 2.8\%. This demonstrates that our proposed KBQG method provides significant assistance to downstream KBQA tasks\footnote{Further analysis will be presented in Appendix~\ref{order_of_demonstration}.}.

\begin{table}[ht]
\centering
\small
\begin{tabular}{cc}
\hline
Method    & F1-Score \\ \hline
RnG-KBQA  & 75.6     \\ \hline
RnG-KBQA+ & 78.4     \\ \hline
\end{tabular}
\caption{\label{kbqg-improve-kbqa-result}
 The result of our approach in improving the performance of KBQA methods.}
\end{table}

\section{Conclusion}
 In this paper, we presented the KQG-CoT approach to tackle few-shot KBQG tasks. 
 KQG-CoT retrieves relevant logical forms from unlabeled data and incorporates their characteristics. It then generates explicit prompt to showcase the reasoning process for complex question generation based on the selected examples.
 Experimental results demonstrate that our approach achieves state-of-the-art performance compared to baselines and even shows competitive results to full-training methods.
 
\section*{Limitations}
Our proposed prompting method, KQG-CoT, partially relies on handcrafted prompts when writing the subquestions. 
However, handcrafted prompts are usually based on the personal knowledge and experience of the exports, which can introduce subjective biases. 
% These factors may result in the model lacking a comprehensive and accurate understanding of the input, as well as being influenced by potential biases. \yscomment{Please remove the url for all the references.}
%Another constraint is that our method is designed for structured input, and may not be as suitable for unstructured input data.

\section*{Acknowledgements}
This work was supported by Natural Science Foundation of China (Project No. 62206097) and Shanghai Pujiang Talent Program (Project No. 22PJ1403000). We sincerely thank the anonymous reviewers for their valuable comments and feedback.

\bibliography{custom}
\bibliographystyle{acl_natbib}

% \newpage
% \input{empty}

% \newpage
\appendix

% \onecolumn
\section{Appendix}

\subsection{Ablation Study on More Datasets}
\label{ablation_study_wq_pq}

We display Table~\ref{main-ablation-wq-pq-result} to show more ablation studies on WQ and PQ datasets.
We can also recognize the significance of our CoT reasoning chain, K-means algorithm, and structure encoding.

\subsection{Illustrative Examples of KQG-CoT+ Prompt}

We present a selection of illustrative examples showcasing our proposed prompts and predictions on WQ, GQ, and PQ  in Table~\ref{main-wq-prompts}, Table~\ref{main-gq-prompts} and Table~\ref{main-pq-prompts}, respectively. As
% \yscomment{please always keep a space between two sentences} each dataset has its own style, it can be observed from the given examples that the prompt style we have constructed aligns well with the style of the dataset.

\subsection{Detailed Prompt Design of KQG-CoT+}
\label{instruction_prompt}
To enhance the guidance provided to LLM in question generation, we have included a descriptive sentence in the demonstrations, which states:
``\textit{Let's engage in a step-by-step exercise of generating questions from logical forms.
% ~\yscomment{please rearrange the instruction with this format}. 
We have provided several examples, each comprising an 'Input' logical form and a corresponding 'Subquestion' that we aim to generate. 
By deconstructing the input logical form into basic components, we can generate questions iteratively until we get the final question.
For each 'Subgraph', we can construct a relevant 'Subquestion' phrase to assist in generating the subsequent question in the sequence.''}.

% \subsection{Effect of Different Chains of Question Generation}
% \label{importance_subquestions}

% We have discovered that the inclusion of manually crafted chains of question generation plays a crucial role in achieving favorable outcomes in our approach. 
% We have extensively explored various techniques for generating these intermediate subquestions and have concluded a chain that prioritizes incorporating substantial information from the subgraph, even at the expense of strict grammatical correctness, yields the most optimal results.

% We believe that the challenge encountered in all approaches that rely on manually constructing intermediate reasoning steps in CoT is a shared one. 
% Therefore, an essential objective for future research is to automate the generation of effective intermediate reasoning steps. 
% 4This research direction is centered on developing methods that can automatically construct intermediate reasoning steps with superior performance.

\subsection{Effect of Demonstration Order}
\label{order_of_demonstration}

During the experiment, we made a noteworthy observation regarding the impact of demonstration order on the performance of our method. 
We conducted a comprehensive exploration of various sorting techniques, including uncertainty-based sorting~\cite{diao-2023:active}, random sorting, and sorting based on the number of logical form jumps.
% ~\yscomment{I suggest you to cite these papers}. 
The detailed experimental results are presented in Table~\ref{table:order_of_demonstration}. 
It becomes evident that arranging the demonstrations in ascending order of the number of logical form jumps leads to the most favorable outcomes. 
This finding highlights the structural complexity of logical forms when organizing the demonstrations.

% During the experiment, we discovered a significant sensitivity of our method's performance to the order of demonstrations. We explored several sorting techniques, including uncertainty-based sorting, random sorting, and sorting based on the number of logical form jumps. The detailed experimental results are presented in Table~\ref{table:order_of_demonstration}. 
% It is apparent that sorting the demonstrations in ascending order of the number of logical form jumps yields the most favorable outcomes.

\begin{table}[ht]
\centering 
\centering  
\small
\begin{tabular}{l | c c c }
\toprule
\multirow{2}{*}{Method} &
  \multicolumn{3}{c}{GQ} \\
 & B & M & R  \\ 
  \midrule
KQG-CoT+ & 31.24 & 34.94 & 45.36 \\
\midrule 
(a) -Uncertainty & 30.36 & 33.91 & 45.05\\
(b) -Similarity & 31.20 & 34.63 & 45.28 \\
(c) -Random & 30.81 &34.26 & 44.91\\
(d) -L2s & 30.52 & 33.66 & 44.83 \\

\bottomrule
\end{tabular}
\caption{\label{table:order_of_demonstration}
 The results of using different sorting methods for demonstrations on the GQ dataset are as follows: Our KQG-CoT+ method is sorted in ascending order of the number of logical form jumps. \textbf{Random} sorting is done randomly. \textbf{L2S} sorting is performed in ascending order of length. \textbf{Uncertainty} sorting is based on descending order of uncertainty values. Lastly, \textbf{similarity} sorting is based on descending order of similarity values between the logical forms of demonstrations and tests.
}
\end{table}

\begin{table*}[ht!]
\centering
\small
\begin{tabular}{l | c c c | c c c}
\toprule
\multirow{2}{*}{Method} &
  \multicolumn{3}{c}{WQ } &
  \multicolumn{3}{c}{PQ}\\
 & B & M & R  & B & M & R \\ 
  \midrule
KQG-CoT+ & 29.73 &31.08 & 55.14  &61.71 &44.27 &78.41\\
\midrule 
(a) w/o CoT & 28.75& 30.12& 54.24 & 60.83 & 44.06 & 77.88 \\
(b) K-means $\rightarrow$ Random & 25.02 & 29.37 & 53.16 & 56.42 & 42.61 & 77.03\\
(c) w/o structure encoding & 28.52 &29.73 & 54.28 &60.34 & 43.26 & 77.59 \\
\bottomrule
\end{tabular}
\caption{\label{main-ablation-wq-pq-result}
 Ablation study of our KQG-CoT+ method on WQ and PQ.
}
\end{table*}

\begin{table*}[ht!]
\centering  
\small
\scalebox{0.96}{
\begin{tabular}{l }
\toprule
Demonstrations \\
\hline
Input: (JOIN (R location.country.official\_language) (JOIN location.country.languages\_spoken romansh language)) \\
Subgraph1: (JOIN location.country.languages\_spoken romansh language)                   \\
Subgraph2: (JOIN (R location.country.official\_language) Subgraph1)                     \\
Subquestion1: country languages spoken romansh language                                 \\
Subquestion2: What is the main language spoken in the country that romansh language is used ?                    \\
\\
        ... \\
        \\
Input: (AND (JOIN people.cause\_of\_death.parent\_cause\_of\_death drug) (JOIN (R people.deceased\_person.cause\_of\_death)   \\ \qquad \quad
(JOIN film.actor.film (JOIN film.performance.character julia biggs)))) \\
Subgraph1: (JOIN people.cause\_of\_death.parent\_cause\_of\_death drug)                                                 \\
Subgraph2: (JOIN film.performance.character julia biggs)                                                                \\
Subgraph3: (JOIN film.actor.film Subgraph2)                                                                             \\
Subgraph4: (JOIN (R people.deceased\_person.cause\_of\_death) Subgraph3)                                                \\
Subgraph5: (AND Subgraph1 Subgraph4)                                                                                    \\
Subquestion1: parent cause of death drug                                                                                \\
Subquestion2: performance character julia biggs                                                                         \\
Subquestion3: film actor who performance julia biggs                                                                    \\
Subquestion4: cause of death of film actor who performance julia biggs                                                  \\
Subquestion5: Which drugs caused the death of the actor who played julia biggs ?                                        \\
\\
\begin{tabular}[c]{@{}l@{}}Input: (JOIN (R location.country.currency\_used) (JOIN location.country.national\_anthem (JOIN \\ \qquad \quad       government.national\_anthem\_of\_a\_country.anthem aruba dushi tera)))\end{tabular} \\
\midrule
Prediction \\
\midrule
Input: (JOIN (R film.performance.actor) (AND (JOIN film.performance.character simon birch) (JOIN film.film.starring (JOIN \\ \qquad \quad 
film.performance.actor ian michael smith)))) \\
Subgraph1: (JOIN film.performance.character simon birch)                                \\
Subgraph2: (JOIN film.performance.actor ian michael smith)                              \\
Subgraph3: (JOIN film.film.starring Subgraph2)                                          \\
Subgraph4: (AND Subgraph1 Subgraph3)                                                    \\
Subgraph5: (JOIN (R film.performance.actor) Subgraph4)                                  \\
Subquestion1: performance character simon birch                                         \\
Subquestion2: performance actor ian michael smith                                       \\
Subquestion3: performance actor ian michael smith star in                               \\
Subquestion4: performance character simon birch the film that ian michael smith star in \\
Subquestion5: Who plays simon birch in the movie that ian michael smith acted in ?     \\
\bottomrule
\end{tabular}
}
\caption{\label{main-wq-prompts}
Prompt with demonstrations and prediction on WQ, the preceding section displays the prompt, and followed section displays the outputs generated by LLMs.}
\end{table*}

\begin{table*}[]
\centering  
\small
\scalebox{0.96}{
\begin{tabular}{l }
\toprule
Demonstrations \\
\hline
Input: (ARGMIN base.exoplanetology.exoplanet astronomy.astronomical\_discovery.discovery\_date)     \\
Subgraph1: (ARGMIN base.exoplanetology.exoplanet astronomy.astronomical\_discovery.discovery\_date) \\
Subquestion1: Which exoplanet was first to be found ?  \\
\\
        ... \\
        \\
Input: (AND digicams.digital\_camera (AND (lt digicams.digital\_camera.weight 250.0\textasciicircum{}\textasciicircum{}http://www.w3.org/2001/XMLSchema \\ \qquad \quad
\#float)(JOIN (R digicams.camera\_viewfinder\_type.digital\_cameras) (JOIN digicams.camera\_viewfinder\_type.digital \\ \qquad \quad
\_cameras Sony Alpha 700)))) \\
Subgraph1: (lt digicams.digital\_camera.weight 250.0\textasciicircum{}\textasciicircum{}http://www.w3.org/2001/XMLSchema\#float) \\
Subgraph2: (JOIN digicams.camera\_viewfinder\_type.digital\_cameras Sony Alpha 700 )                                             \\
Subgraph3: (JOIN (R digicams.camera\_viewfinder\_type.digital\_cameras) Subgraph2)                                               \\
Subgraph4: (AND Subgraph1 Subgraph3)                                                                                             \\
Subgraph5: (AND digicams.digital\_camera Subgraph4)                                                                              \\
Subquestion1: digital cameras that weight less than 250.0                                                                        \\
Subquestion2: viewfinder type digital cameras sony alpha 700                                                                     \\
Subquestion3: digital cameras the same viewfinder type as the sony alpha 700                                                     \\
Subquestion4: digital cameras the same viewfinder type as the sony alpha 700 and weight less than 250.0                          \\
Subquestion5: Are there any digital cameras that use the same viewfinder as the sony alpha 700 that weight less than 250.0?      \\
\\
Input: (AND music.genre (JOIN (R music.genre.parent\_genre) (JOIN music.genre.albums confessions tour)))  \\
\midrule
Prediction \\
\midrule
Subgraph1: (JOIN music.genre.albums confessions tour)                                                    \\
Subgraph2: (JOIN (R music.genre.parent\_genre) Subgraph1)                                               \\
Subgraph3: (AND music.genre Subgraph2)                                                                  \\
Subquestion1: the music genre albums confessions tour                                                     \\
Subquestion2: the albums confessions tour is part of what parent genre                                   \\
Subquestion3: The albums confessions tour is part of what parent genre of a musical genre?                  \\ 
\bottomrule
\end{tabular}
}
\caption{\label{main-gq-prompts}
 Prompt and prediction on GQ, the preceding section is the prompt, and the blue text following it represents the prediction.
}
\end{table*}

\begin{table*}[]
\centering  
\small
\scalebox{0.96}{
\begin{tabular}{l}
\toprule
Demonstrations \\
\hline
Input: (JOIN (R people.person.gender) (JOIN (R people.person.parents) sviatoslav ii of kiev))                                              \\
Subgraph1: (JOIN (R people.person.parents) sviatoslav ii of kiev)        \\
Subgraph2: (JOIN (R people.person.gender) Subgraph1)                     \\
Subquestion1: sviatoslav ii of kiev 's parents                           \\
Subquestion2: What is the gender of sviatoslav ii of kiev 's dad ?       \\
\\
        ... \\
        \\
Input: (JOIN (R people.deceased\_person.place\_of\_death) (JOIN (R people.person.children) (JOIN (R people.person.children) p \\
\qquad \quad   j kennedy))) \\
Subgraph1: (JOIN (R people.person.children) p j kennedy)                 \\
Subgraph2: (JOIN (R people.person.children) Subgraph1)                   \\
Subgraph3: (JOIN (R people.deceased\_person.place\_of\_death) Subgraph2) \\
Subquestion1: p j kennedy 's children                                    \\
Subquestion2: children of p j kennedy 's children                        \\
Subquestion3: What is the place of death of kid of p j kennedy 's son ? \\
\\
Input: (JOIN (R music.recording.releases) (JOIN (R music.recording.tracks) o holy night))                 \\
\midrule
Prediction \\
\midrule
Subgraph1: (JOIN (R music.recording.tracks) o holy night)                                                \\
Subgraph2: (JOIN (R music.recording.releases) Subgraph1)                                                \\
Subquestion1: o holy night 's tracks                                                                     \\
Subquestion2: What is the releases of recording of o holy night 's tracks ?  
\\
\bottomrule
\end{tabular}
}
\caption{\label{main-pq-prompts}
 Prompt and prediction on PQ, the preceding section is the prompt, and the blue text following it represents the prediction.
}
\end{table*}

\label{sec:appendix}

\end{document}